\documentclass[letterpaper, 10 pt, journal, twoside]{IEEEtran}

\usepackage{graphicx}
\usepackage{mathptmx}
\usepackage{times}
\usepackage{amsmath}
\usepackage{amssymb}
\usepackage{glossaries}
\usepackage{microtype}
\usepackage{hyperref}

\title{\LARGE \bf
Rewarding DINO: Predicting Dense Rewards\\ with Vision Foundation Models
}

\author{
    Pierre Krack$^{1}$, Tobias Jülg$^{1}$, Wolfram Burgard$^{1}$ and Florian Walter$^{1,2}$%
\thanks{$^{1}$Department of Computer Science~\& Artificial Intelligence, University of Technology Nuremberg, Germany. Contact: \texttt{pierre.krack@utn.de}}%
\thanks{$^{2}$TUM School of Computation, Information and Technology, Technical University of Munich, Germany}
}

\newcommand{\etal}{et\penalty50\ al.\@}
\newcommand{\sgn}{\mathop{\mathrm{sgn}}}
\newcommand{\method}{Rewarding DINO}

\setacronymstyle{long-short}
\newacronym{rl}{RL}{Reinforcement Learning}
\newacronym{irl}{IRL}{Inverse Reinforcement Learning}
\newacronym{mlp}{MLP}{multi-layer perceptron}
\newacronym{vit}{ViT}{Vision Transformer}
\newacronym{bert}{BERT}{bidirectional encoder representations from transformers}
\newacronym{sbert}{SBERT}{sentence BERT}
\newacronym{llm}{LLM}{Large Language Model}
\newacronym{vlm}{VLM}{Vision-Language Model}
\newacronym{film}{FiLM}{feature-wise linear modulation}
\newacronym{ece}{ECE}{estimated calibration error}
\newacronym{ppo}{PPO}{proximal policy optimization}
\newacronym{sb3}{SB3}{Stable-Baselines3}
\newacronym{pbrs}{PBRS}{potential-based reward shaping}

\begin{document}

\maketitle
\thispagestyle{empty}
\pagestyle{empty}

\begin{abstract}
    Well-designed dense reward functions in robot manipulation not only indicate whether a task is completed but also encode progress along the way.
    Generally, designing dense rewards is challenging and usually requires access to privileged state information available only in simulation, not in real-world experiments.
    This makes reward prediction models that infer task state information from camera images attractive.
    A common approach is to predict rewards from expert demonstrations based on visual similarity or sequential frame ordering.
    However, this biases the resulting reward function towards a specific solution and leaves it undefined in states not covered by the demonstrations.
    In this work, we introduce \method{}, a method for language-conditioned reward modeling that learns actual reward functions rather than specific trajectories.
    The model’s compact size allows it to serve as a direct replacement for analytical reward functions with comparatively low computational overhead.
    We train our model on data sampled from 24 Meta-World+ tasks using a rank-based loss and evaluate pairwise accuracy, rank correlation, and calibration.
    \method{} achieves competitive performance in tasks from the training set and generalizes to new settings in simulation and the real world, indicating that it learns task semantics.
    We also test the model with off-the-shelf reinforcement learning algorithms to solve tasks from our Meta-World+ training set. 
\end{abstract}

\section{Introduction}
Performance metrics are essential for learning and evaluating robot manipulation policies.
In reinforcement learning, these metrics are provided by reward functions that map a robot’s task performance to a simple numerical value.
However, defining dense reward functions for robotic tasks that encode fine-grained feedback is challenging, and often depends on privileged state information available only in simulation.
Modern vision foundation models can encode both high-level semantics and fine-grained visual details, suggesting that their representation may contain the information necessary to compute such rewards.
We ask the evident question: can dense reward functions be predicted from pre-trained image and language embeddings, with a precision comparable to rewards computed from privileged state?

Reward prediction is appealing due to its many downstream applications: policy evaluation, dataset analysis, AI safety, and reinforcement learning.
The task naturally spans a broad range of difficulty and is hardware agnostic, enabling researchers to flexibly scale data and model capacity.
These properties make it a promising candidate for large-scale training.

Reward prediction has recently garnered significant attention~\cite{yang_rank2reward_2024,liang2026robometer,lee2026roboreward,zhang2025rewind,fan_minedojo_2022,ma_vip_2022,ma_liv_2023}.
Most methods leverage successful task demonstrations to implicitly define rewards through visual similarity or by analyzing the temporal order of frames in demonstration videos.
The main focus is on scaling the methods to large datasets.
However, unlike traditional analytic reward functions, this ties the resulting reward prediction closely to a specific solution rather than to the task definition itself.
In particular, the resulting reward models are only defined within the state space defined by the collected demonstrations.
This is problematic when training a reinforcement learning policy from scratch, as the trajectories during early training phases may differ substantially from expert demonstrations.

\begin{figure}
\centering
\includegraphics{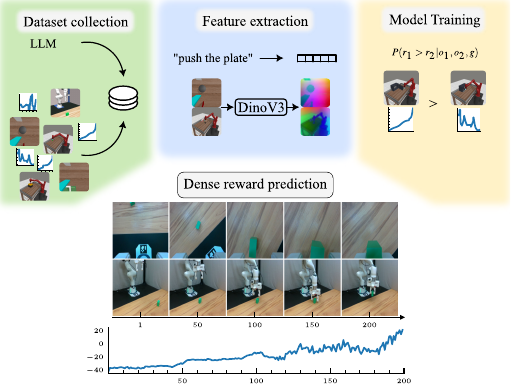}
\caption{We collect a dataset with multiple camera views and associated rewards across multiple tasks, formulate reward modeling as a pairwise preference learning problem, and recover dense shaped reward functions.}
\end{figure}

In this work, we introduce \method{}, a method that addresses these limitations by learning dense rewards that can be computed analytically.
This sidesteps the lack of annotated non-expert data, allowing us to answer whether vision foundation models can and should serve as a backbone for reward prediction.
Our approach enables evaluation with respect to analytical rewards and can be used to solve Meta-World+ tasks from scratch with off-the-shelf reinforcement learning algorithms.
Importantly, our formulation does not sacrifice generality, as it naturally encompasses alternative supervision signals---including sparse rewards, preference data, or demonstration data.
Our approach can therefore be viewed as a recipe for training precise dense reward models.

\textbf{In summary, we make the following contributions:}
\begin{enumerate}
\item We formulate reward prediction as a supervised learning problem applicable to several existing approaches.
\pagebreak
\item We design and train a lightweight architecture to predict dense rewards from encodings of pre-trained vision and language models.
\item We evaluate the precision, calibration, and generalization capabilities of our approach.
\item We show a principled method for training policies \emph{from scratch} using image-based reward models.
\end{enumerate}

\section{Related Work}
Our work is related to \gls{irl}~\cite{ng_algorithms_2000} in that we also learn a reward function.
\gls{irl} uses unlabeled demonstrations, whereas we rely on existing reward functions to generate data.
Within this scope, the work closest to ours is that of Fu~\etal{}~\cite{fu2018from}, who introduce language conditioning to learn goal-conditioned reward functions. 

More recent work focuses on predicting rewards from videos with contrastive learning.
MineDojo~\cite{fan_minedojo_2022}, VIP~\cite{ma_vip_2022}, and LIV~\cite{ma_liv_2023} use a CLIP-like objective~\cite{radford2021learning} to align images of solved tasks with corresponding textual goal descriptions.
In particular, VIP and LIV focus on pre-trained representations for robot policies and show that they can be interpreted as progress indicators.
The limitation of such an approach is that visual similarity cannot always be equated with task progress.

Another line of work has investigated the use of \glspl{vlm} for success detection.
In RoboFume~\cite{yang_robot_2023}, the authors use a \gls{vlm} fine-tuned on task success detection to fine-tune pre-trained policies on novel tasks.
Similarly, Yuqing~\etal{} fine-tune \glspl{vlm} to reliably detect success and evaluate the precision of success detection~\cite{vlm_success_detectors}.
Different from these works, \method{} learns a dense reward function.

\glspl{llm} and \glspl{vlm} have also been used to generate code for reward functions. 
Yu~\etal{}~\cite{yu_language_2023} generate residuals for optimization using model predictive control.
Kwon~\etal{}~\cite{kwon_reward_2022} utilize an \gls{llm} directly inside a \gls{rl} loop as a reward function.
Eureka~\cite{ma_eureka_2024} is also based on an \gls{llm}, but uses it to iteratively improve its reward function code after observing policies trained with the generated reward. Instead, 
\method{} does not generate code but directly maps observations to scalar rewards.

Several recent works focus on reward learning, leveraging large demonstration datasets and the order of frames within demonstration videos to recover a reward signal. 
RankNet~\cite{yang_rank2reward_2024} introduces the idea and employs an adversarial algorithm to train models with the learned rewards. 
ReWiND~\cite{lee2026roboreward} proposes video rewinding to learn reverse progress, and fine-tunes pretrained policies on new tasks where the initial policy performs adequately.
RoboMeter~\cite{liang2026robometer} attempts to further mitigate the lack of negative examples by including an additional prediction head, which is trained separately to prefer expert demonstrations to failed or suboptimal trajectories.
They then train an SAC policy to shape the noise of a pre-trained diffusion model using DSRL~\cite{wagenmaker2025steering}.
The authors of RoboReward~\cite{lee2026roboreward} present a dataset and a benchmark, where partially and unsolved tasks are generated via counterfactual prompts generated by a \gls{vlm}.
Both augmentations also address the lack of suboptimal and failed trajectories.

We focus on developing a simple, data-efficient, and effective supervised formulation for learning well-defined, dense rewards from pretrained vision and language representations, without sacrificing generality. 
This formulation allows us to directly evaluate reward prediction against an analytical ground truth and to measure how this translates into training \gls{rl} policies.

\begin{figure}
\centering
    \includegraphics[width=\linewidth]{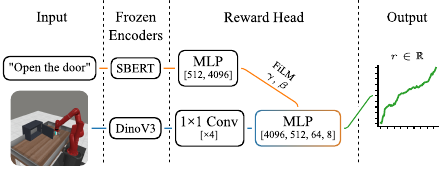}
    \caption{
        Model architecture overview.
        The text and image encoders are frozen.
        We train only the small head, which fuses language and image tokens through \gls{film}.
        The inputs are two 512$\times$512 images and a textual goal description; the output is a single scalar.
    }
    \label{fig:architecture}
\end{figure}

\section{Method}
This section describes our reward learning method.
Our goal is to map visual observations from a robot manipulation task to a numeric score based on a natural language task description.
While sequential models can be more expressive, we focus on learning simple reward functions that, like typical analytical reward functions, depend only on the current state of the robot and the environment, not on history.
Reward functions can be concatenated by changing the prompt over time.

Given an observation $o_t = (I_1, I_2)$ composed of two image embeddings $I_1, I_2$, and an embedding of a textual goal description $g$, the objective is to learn a function $f(o_t, g)$ that computes a scalar score $s$ inducing the same ordering over states as a ground truth reward $r_t$.
In the following, we motivate and describe our modeling choices and the objective function before describing the dataset collection.

\subsection{Model Architecture}
Figure~\ref{fig:architecture} provides an overview of our architecture.
One of our key ideas is that a strong, dense image representation, combined with a goal embedding, reduces reward learning to measuring distances to and between objects specified by the goal description and verifying discrete conditions related to the environment state.
The model must therefore fuse image embeddings with embeddings of goal descriptions to infer the corresponding reward.
Because this is only possible with a clear and unobstructed view of all task-relevant objects, we use two camera perspectives to reduce occlusions and ambiguities that can arise from single-view observations.

We chose DINOv3~\cite{simeoni2025dinov3} in its variant \textit{ViT-S/16} as our image embedding model, a \gls{vit} with $21\,\mathrm{M}$ parameters trained with a particular emphasis on dense features.
The model's comparatively small size reduces the computational costs for inferencing rewards, which is a key prerequisite for deploying our model as a drop-in replacement for analytical reward functions.
To encode goals, we use \textit{allMiniLM-L6-v2}, a \gls{bert} model~\cite{devlin2019bert} distilled using deep self-attention~\cite{wang2020minilm} and fine-tuned using the approach described in \gls{sbert}~\cite{reimers2019sentence}.

Since we rely on dense spatial information, we use the full patch embedding matrix of DINOV3, a grid of tokens with positional information, each representing a specific part of the image.
We do not use the class token, which represents the whole image in a single token.
Moreover, we use four one-by-one convolution kernels to project each 384-dimensional token into a 4-dimensional vector.
This is equivalent to having four separate linear layers, each applied to each token individually.
The result is concatenated into a vector and fed into a small \gls{mlp}---the reward head.

The reward head consists of four linear layers with sizes 4096, 512, 64, and 8.
Each of them is followed by layer normalization~\cite{ba2016layer}, and we use LeakyReLU~\cite{maas2013rectifier} activation functions.
The first three layers merge the goal description using \gls{film}~\cite{film}.

\gls{film} is a simple alternative to cross-attention for conditioning.
Both \gls{film} and cross-attention learn parameters for a parameterized function.
Whereas cross-attention computes matrices to be used in a matrix multiplication, \gls{film} computes a vector to be used in a simpler affine transformation \(\gamma \odot x + \beta\), where $\odot$ represents the Hadamard product, and $\gamma$ and $\beta$ are the \gls{film} parameters.
\gls{film} rescales and shifts the individual feature dimensions, and gains representational power from nonlinearities in consecutive modulated layers.
The \gls{film} parameters are computed by a \textit{\gls{film} generator}.
In our case, the generator is another small \gls{mlp} which takes the goal embedding as input, and outputs the parameters $\gamma$ and $\beta$ for each of the first three layers of the reward head.

Our model maps each image observation deterministically to a single scalar score $s$.
A consequence of this design is that the outputs of our model, when interpreted as a scoring function, induce a total pre-order over observations, provided the goal embedding does not change.

\subsection{Pairwise Ranking Loss}
\label{subsec:method-loss}
Rewards can be interpreted as a numerical representation of preferences over trajectories, as posited by the reward hypothesis~\cite{sutton2020reinforcement} and formalized in Bowling \etal{}~\cite{bowling2023settling}.
As shown by Ng \etal{}~\cite{ng1999policy}, optimal policies are invariant to positive affine transformations of the reward function, implying many different reward functions can induce the same preference relationship over trajectories.
In short, the absolute scale of rewards is not generally meaningful.
Instead, the ordering they induce matters more.

We reflect this in our training objective by using a pairwise logistic loss, as introduced in RankNet~\cite{burges2005learning}, rather than regressing absolute reward values.
Let 
\begin{equation}
y =
\left\{
\begin{array}{ll}
\phantom{-}1  & \textrm{if } r_0 > r_1 \\
-1 & \textrm{if } r_1 > r_0
\end{array}
\right.
\end{equation}
be a binary classification label and $\Delta s = s_0-s_1$ the difference between the predicted scores of the first and second sample.
The logistic pairwise loss can be written as
\begin{equation}
    \ell(\Delta s) = \log\left(1 + \exp\left(\frac{-y\Delta s}{\tau}\right)\right),
\end{equation}
where $r_0$ and $r_1$ are the ground truth rewards, and $\tau$ is a temperature parameter controlling the sharpness of the loss.
Lower values of $\tau$ result in higher errors for mis-ordered pairs, which then decay faster once the pair is ordered correctly, and vice-versa.
Pairs with equal rewards do not express a strict preference, so we exclude them from training.
This avoids imposing arbitrary equality constraints on predicted scores for indifferent pairs.
The pairwise formulation allows us to use arbitrary reward scales across tasks, train with binary success indicators, or use the order of frames in demonstrations.

Note that the pairwise logistic loss is conceptually similar to contrastive losses, as introduced by Hadsell \etal{}~\cite{hadsell2006dimensionality}, and shares many of their advantages.
It focuses optimization on ambiguous or misordered pairs while avoiding the need to predict potentially irrelevant absolute reward values.

\subsection{Data Generation and Sampling}
\begin{figure}
\centering
\includegraphics[width=0.9\linewidth]{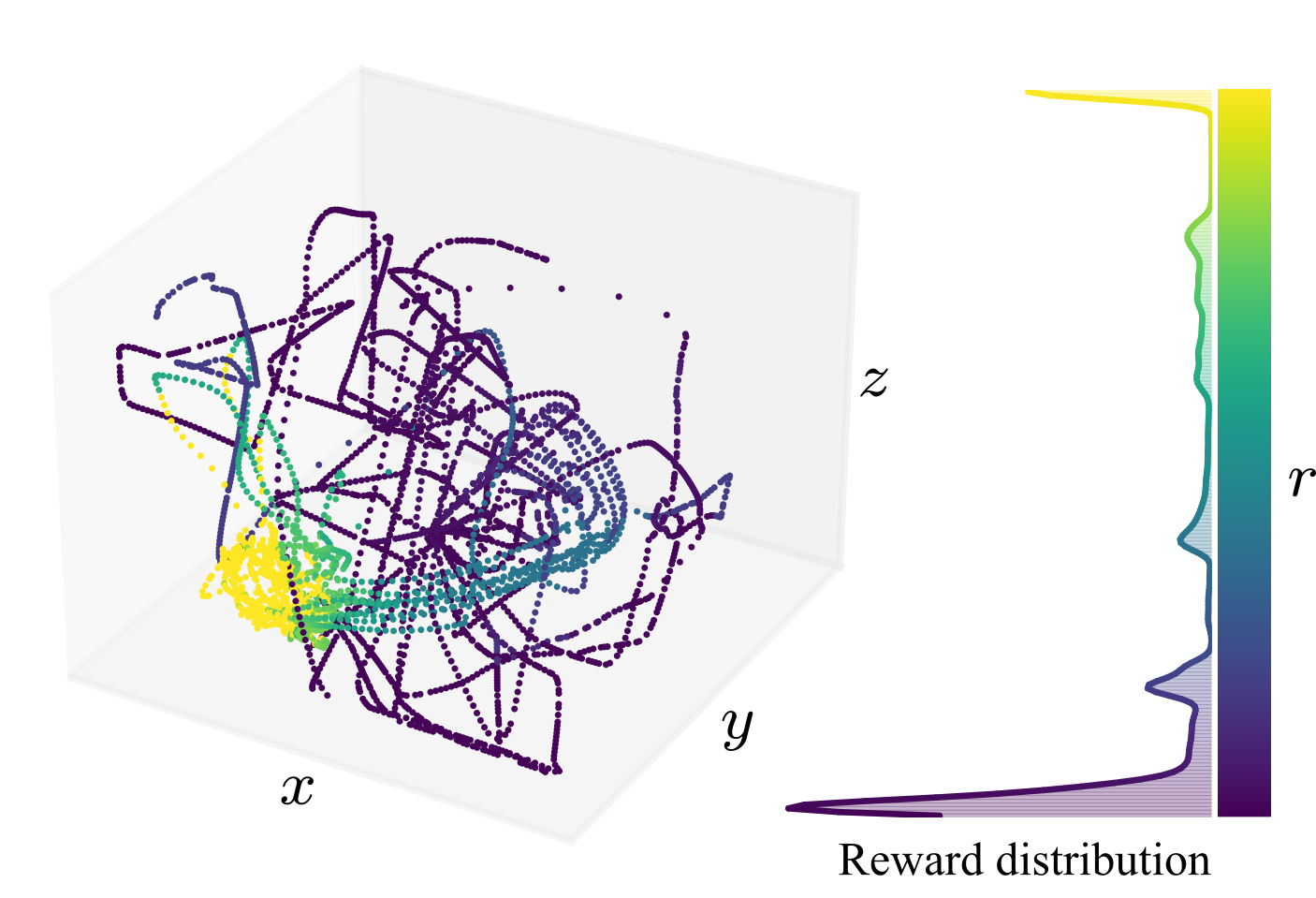}
\caption{Plot of the TCP Cartesian coordinates in the training dataset for the door-open task after binning.
The histogram on the right shows the distribution of rewards in the sample.
}
\label{fig:data_dist}
\end{figure}

\label{sec:method-data}
The goal of our method is to learn reward prediction based on existing expert rewards.
For data collection, we thus assume that a task environment with an appropriate reward function is given.
In robotics, collecting a comprehensive dataset is difficult in itself.
Collecting random data often fails to cover a large state space, whereas using expert policies typically yields trajectories that are similar.
We therefore collect both random and expert trajectories, where the expert policies switch to a random policy when the task is solved and back to expert when it becomes unsolved.
Note that a naive random policy is a poor data-collection strategy, as the robot mostly remains in one place, depending on how actions are defined.
In our case, we pick a random action and repeat it $n$ times to cover a larger space.

Since we collect trajectories and associated step-wise rewards, we can verify the coverage in both Cartesian and reward space.
If rewards correspond to progress along tasks, covering a broad spectrum of rewards implies coverage of task-relevant states.
A good coverage in Cartesian space does not imply a good coverage in task space and vice versa.

Even after ensuring broad coverage of both rewards and Cartesian coordinates, the dataset remains unbalanced, as can be seen in the reward histogram for an example task in Figure~\ref{fig:data_dist}).
Despite randomization, there are many samples corresponding to the initial positions of the task and, accordingly, low rewards.
Analogously, there are many samples where the task is solved and, accordingly, high rewards.
The combination of reward and Cartesian coordinates provides insights into which samples are unique.
Based on this insight, we partition the dataset into unique bins of reward and Cartesian coordinates and then sample one element from each bin.
This deduplication strategy, which removes samples where both the reward and the Cartesian coordinates differ by at most 0.01, roughly halves the number of samples.
Figure~\ref{fig:data_dist} depicts the retained samples from a single task after binning.

\section{Results}
To evaluate our model, we consider different metrics to quantify the quality of the rewards predicted by the model relative to analytical ground-truth rewards on separately collected trajectories, and show the performance gap between random and expert trajectories.
As reward prediction from images will degrade, for example, if a scene is occluded, we show a natural probabilistic interpretation of our model that allows for quantifying the confidence of the predicted rewards.
We further investigate generalization to new task variations (e.g.\@ open vs.\@ close an object) and visual domain shift (simulated vs.\@ real-world experiments).
To analyze whether our model can replace analytical reward functions, we also train policies with \gls{ppo}~\cite{schulman2017proximal} across multiple tasks and compare them against the ground-truth dense reward and sparse rewards.

\subsection{Experiment Setup}
In the scope of this work, we focus on manipulation tasks centered around a robot with a gripper that manipulates objects on a table.
We capture multiple images per sample, one of which is always from a gripper-mounted camera.

We use Meta-World+~\cite{mclean2025metaworld} and a simulated pick-cube replica of our lab setup to collect data for our experiments.
Our lab setup consists of a Franka Research 3 robot mounted on a table.
A RealSense camera is attached to the gripper, and 4 other RealSense cameras show the scene from different angles.
We simulate the scene in MuJoCo~\cite{todorov2021mujoco} using Robot Control Stack~\cite{rcs} and use a ManiSkill-inspired~\cite{mu2021maniskill} reward function.

The time interval between consecutive samples significantly affects the difficulty.
In our case, our model is trained to distinguish rewards from states that are 1/80~s apart in the Meta-World+ tasks and 1/30~s apart in the pick cube task.

\subsubsection{Dataset}
\begin{figure}
    \includegraphics[width=\columnwidth]{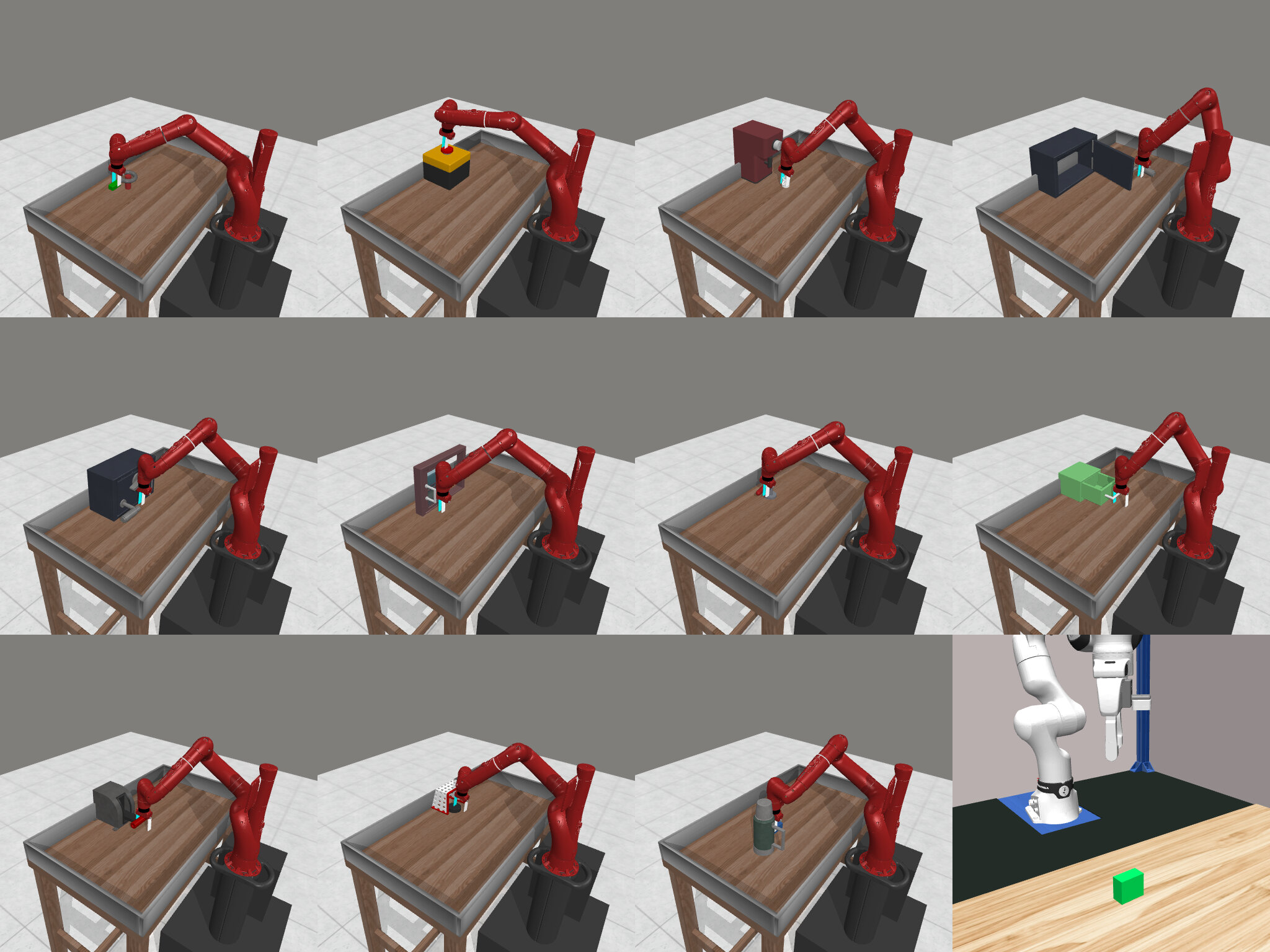}
    \caption{
    All the scenes and objects encountered in the training set.
    The objects move around, and the tasks have variations.
    The training tasks are: assembly, disassemble, button-press-topdown, door-open, door-close, window-close, window-open, drawer-open, faucet-open, faucet-close, handle-press, handle-pull, handle-press-side, handle-pull-side, door-lock, door-unlock, stick-pull, stick-push, plate-slide, plate-slide-side, plate-slide-back, plate-slide-back-side, coffee-pull, coffee-push, and our simulated pick task.
    }
    \label{fig:data_all}
\end{figure}

We use 24 of the Meta-World+ environments, selecting those that include variations (e.g., opening versus closing a door) so that the model must rely on language instructions.
We keep drawer-close and its mostly binary rewards as a separate test task, to investigate the model's generalization capabilities.
Figure~\ref{fig:data_all} shows all the tasks in our dataset.

The dataset collection requires only a single hyperparameter $n$ to specify the number of times an action is repeated.
This is necessary because it is task- and environment-dependent.
Sampling is parameterized by $\epsilon_c$ and $\epsilon_r$, which determine the bin resolution for the cartesian positions and the rewards, respectively.
After normalizing rewards, we use them to identify duplicate samples.
In our case, if the Cartesian distance between two samples is less than a centimeter ($\epsilon_c = 0.01$) and the rewards differ by less than 1\% of the total reward variation within the task ($\epsilon_r = 0.01$), we consider the samples duplicates.

In total, our dataset consists of 770 trajectories, totaling 169,018 distinct steps after binning, and 809,886 images.
Our sampling procedure draws from more than 900 million step pairs and more than 4 billion image pairs.

\subsubsection{Model Training}
During training, the position of the gripper-mounted camera is randomized, and the second camera view is sampled from multiple alternatives.
We sample only pairs with unequal rewards, since our loss is not defined for pairs with equal rewards.
We also sample the language prompts for each task from a pre-defined set.

We use the Adam optimizer~\cite{adamw} with decoupled weight decay, introducing hyperparameters $\beta_1$ and $\beta_2$ that we set to their default values, along with a weight decay hyperparameter.
The temperature, $\tau$, in the loss function is perhaps the most important hyperparameter as it determines the loss landscape.

In our experiments, the training turned out to be relatively insensitive to the choice of hyperparameters.
We set the learning rate to $0.0003$, the weight decay to $0.03$, and the loss temperature to $2$.
Furthermore, we use a batch size of 512, which, in our environment, is the maximum that still improves throughput,  and train the model for 3000 epochs, where each epoch consists of 100,000 pairs.

\subsection{Reward Quality Metrics}
We quantify the quality of the rewards predicted by the model using the ground-truth analytical rewards from the task environments. 
Because our model is not trained to regress exact values, a direct numerical comparison of the predictions with the ground truth is not possible.
Instead, we start by measuring the \emph{pairwise accuracy} between observation pairs.

\subsubsection{Pairwise Accuracy}
\begin{figure}
\centering
\includegraphics[width=\linewidth]{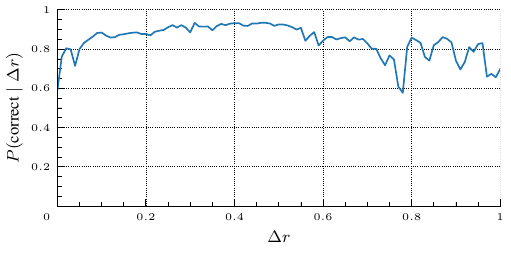}
\caption{
Pairwise accuracy across all training tasks on newly collected trajectories.
We calculate the pairwise accuracy stratified by reward difference.
The y-axis is the likelihood that a pair of samples from the same task but potentially different episodes, using the same prompt and camera configuration, is ranked correctly, given a binned difference in ground-truth rewards on the x-axis.
}
\label{fig:accuracy}
\end{figure}

We measure the pairwise accuracy on observation pairs from the same task, including distinct trajectories, using the same camera perspectives and language prompt, and on a separate dataset collected as described in Section~\ref{sec:method-data}.
We only consider pairs where the normalized rewards differ by at least 0.01 (as in the training procedure) and select the best-performing camera configuration and prompt for each task.
The stratified accuracies are reported in Figure~\ref{fig:accuracy}.
Our model can accurately rank pairs with a small reward difference.
The accuracy exceeds 80\% when the reward difference is between 0.06 and 0.7.
For larger reward differences, accuracy drops, which can be explained by expert policies switching to random as soon as the task is solved.
In such cases, two states can appear visually similar but have drastically different rewards, and our model cannot, for example, determine the exact frame when an object slips from the robot's gripper.

\subsubsection{Rank Correlation}
\begin{figure}
    \centering
    \includegraphics[width=\linewidth]{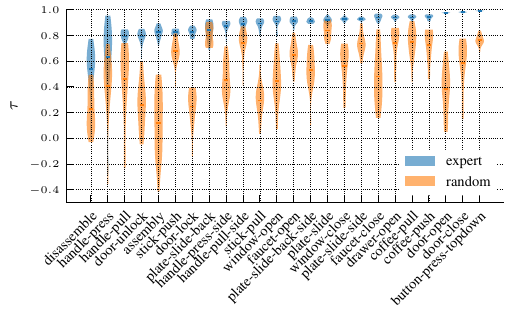}
    \caption{
    Distributions of Kendall's tau ($\tau$) per environment and per policy.
    Kendall's tau is computed on ten expert and ten random trajectories for each environment.
    }
    \label{fig:correlation}
\end{figure}

Next, we measure whether the rank induced by our model's predictions matches the rank defined by the ground-truth rewards.
We use random policies, as described in Section~\ref{sec:method-data}, and expert policies, in which the task finishes as soon as it is solved.
Kendall's tau is a rank-correlation metric with an intuitive interpretation.
It can be written as
\begin{equation}
    \tau = \frac{2}{n(n-1)}\sum_{i < j} \sgn(x_i -x_j) \sgn(y_i-y_j)\textrm{,}
\end{equation}
where $n$ is the number of samples, and $x_k$ and $y_k$ are scalars of a first and a second sequence.
In the absence of ties, a Kendall tau correlation of $-1$ represents a flipped order, $1$ indicates that the order is identical, and $0.5$ indicates that 75\% of the consecutive pairs are ordered correctly.
We use a bias-corrected implementation that considers ties.
This metric is more difficult than pairwise accuracy: Kendall tau only considers consecutive pairs, which will necessarily have closer rewards, especially in random trajectories where the rewards stay low.
Unlike pairwise accuracy, it considers only pairs from the same trajectory.
We report Kendall tau for expert and random trajectories in Figure~\ref{fig:correlation}.
Note the large difference between expert and random trajectories, even though our model is trained on both.
This intuitively makes sense, since the rewards from random trajectories are much lower.
Thus, their order is based on much smaller differences in reward.
Predicting rewards for random trajectories is therefore significantly more difficult, yet crucial for reinforcement learning from scratch, where policies are initially random.


\subsection{Probabilistic Interpretation}
Predicting rewards from images poses several challenges.
Parts of the scene may be occluded or outside the cameras' fields of view, and image resolution dictates the precision with which rewards can be predicted.
As such, it is a natural fit for probabilistic modeling.
With our approach, the distance between two logits naturally leads to a probabilistic interpretation.
The larger the distance, the higher the likelihood that one state is better than another.

\begin{figure}
\centering
\includegraphics{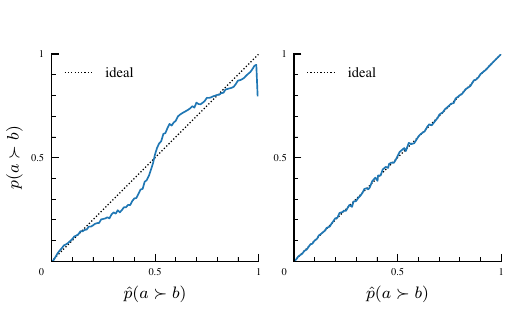}
\caption{
Reliability diagram of the probabilistic interpretation of the model after temperature scaling.
The diagonal line represents a perfectly calibrated model.
When the model is overconfident, the line is below the diagonal and vice versa.
}
\label{fig:calibration}
\end{figure}

We calibrate our model's output based on the difference between two scores with temperature calibration and isotonic regression and report the \gls{ece} between binned pairs in Figure~\ref{fig:calibration}.
The temperature-calibrated model fits a single parameter $\tau$ so that our model prediction $\hat{p}(s_0>s_1) = \sigma((s_0-s_1)/\tau)$ aligns with the empirical likelihood that $p(r_0 > r_1)$.
Isotonic regression, on the other hand, fits a piecewise monotonic transformation of $s_0-s_1$.
Both are calibrated on a train split of the evaluation data, and the plots show the results for a test split.

The temperature-calibrated reward model exhibits an \gls{ece} of 0.043, and the one calibrated with isotonic regression exhibits an \gls{ece} of 0.001.
In other words, when our model assigns a probability of $x\%$ to state $a$ having a higher reward than state $b$, then it is, on average, off by $4.3\%$ or $0.1\%$ respectively.

\subsection{Generalization to New Prompts, Tasks, and Domains}
For our model to be versatile, generalization to settings outside the training set is essential. We analyze generalization with respect to variations in the input prompt, the task, and in a sim-to-real setting.

\subsubsection{Prompt Variation}
We start by investigating the simplest form of generalization: across multiple paraphrases of the prompt.
Given that we already train on multiple prompt variations and use a frozen pre-trained language model to embed goal descriptions, we expect low variance between train and test prompts.
Indeed, pairwise accuracy decreases by 0.002, and per-trajectory Kendall tau decreases by 0.018.

\subsubsection{Task Variation}
Next, we investigate generalization to task variations: can the model predict a dense reward for the drawer-close task, even though it was only shown the drawer-open task during training?
Given the small scale of the dataset, generalization is challenging and requires the model to learn a semantic understanding of the environment and the task based on the representations provided by the pre-trained encoders.
In our experiments, pairwise accuracy drops from 0.93 to 0.51, a decrease of 0.42 points, and Kendall tau drops from 0.82 to 0.47, a decrease of 0.34.
This is due to the ground truth rewards from Meta-World+ being mostly binary for drawer-close and dense for drawer-open.
The effect stems from the random trajectories, in which the ground-truth reward is almost always zero, and the model's predictions are dense.
On expert trajectories, the metrics are substantially better: Kendall's tau drops from 0.94 to 0.72.
Overall, these results highlight the usefulness of the pre-trained encodings for reward prediction.

\subsubsection{Sim-to-Real}
\begin{figure}
\includegraphics[width=\linewidth]{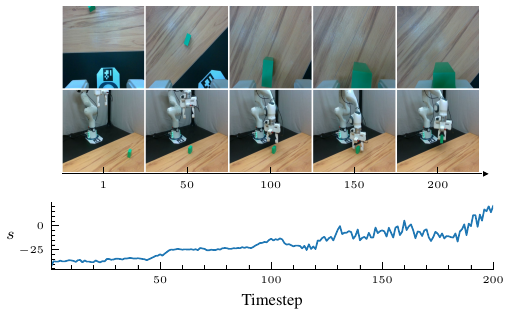}
\caption{Qualitative evaluation of sim-to-real reward prediction. The images correspond to the two camera angles the model sees. The prompt is known from the training set.}
\label{fig:hw}
\end{figure}

\begin{figure*}
    \includegraphics[width=1\columnwidth]{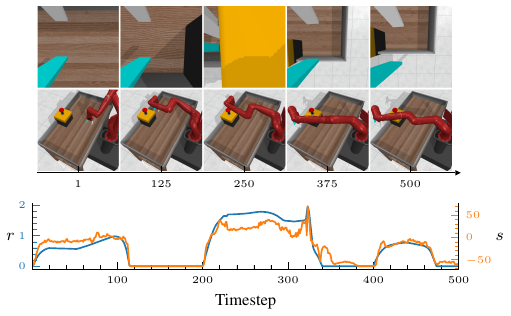}
    \hfill
    \includegraphics[width=1\columnwidth]{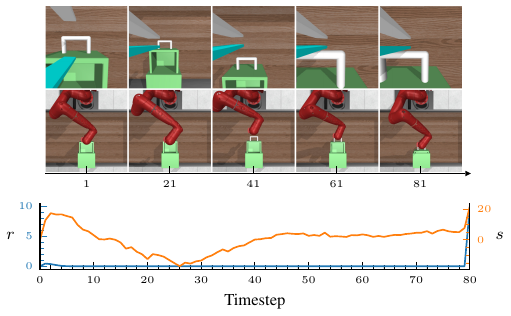}
    \caption{
    Predicted (orange) and ground truth (blue) rewards for a random policy on a known task (button-press-topdown, left) and an expert policy on an unknown task (door-close, right).
    Note that the ground-truth reward for door-close is mostly binary.
    Our model predicts higher rewards at the beginning, which then decrease as the expert policy moves farther away from the drawer (compare $t=1$ with $t=21$), before peaking when the robot closes the drawer.
    A reward model trained on this expert policy would encourage the robot to first move away from the drawer before closing it.
    }
    \label{fig:sim}
\end{figure*}

To analyze the models' sim-to-real performance, we include a pick-cube task simulated in MuJoCo and trained on data collected from different camera angles.
We also record expert trajectories on real hardware and predict their rewards.
Because there is no ground truth for the hardware setup, we present a qualitative evaluation of the reward model on the real setup in Figure~\ref{fig:hw}.
Figure~\ref{fig:sim} shows similar plots for simulated Meta-World+ tasks.
The shaped reward signal computed on the sim-to-real task also demonstrates that our method does not overfit to individual pixels or specific feature dimensions of the encoded image, but instead learns to infer a reward based on relationships among multiple semantically meaningful dimensions in the image embeddings.

\subsection{Reinforcement Learning}
\begin{figure}
    \includegraphics[width=\linewidth]{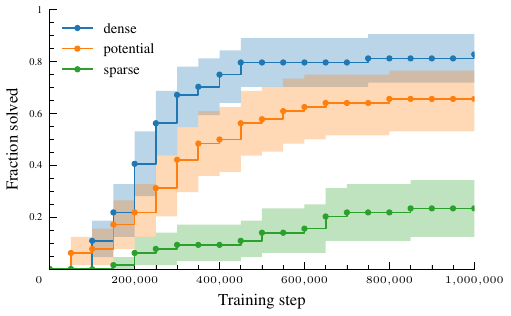}
    \caption{Fraction of runs solved for each reward over the course of one million training steps.
    The shaded area shows 95\% bootstrapped confidence intervals.}
    \label{fig:rl}
\end{figure}
Finally, we train policies from scratch using \gls{ppo} on eight tasks on which the reward model was trained.
The observation and action spaces are unmodified from Meta-World+ (proprioceptive and object state, relative Cartesian TCP actions).
Our selection includes difficult and easy tasks, as well as tasks where our model performs poorly: assembly, button-press-topdown, door-open, drawer-open, faucet-close, plate-slide, window-close, and window-open.
We train \gls{ppo} for at most one million steps and stop training once the policy achieves a 100\% success rate across ten evaluation rollouts, during which actions are deterministic.
We use \gls{sb3}'s~\cite{raffin2021stable} default \gls{ppo} hyperparameters.

Instead of replacing the rewards with our model's reward, we use binary success indicators as the base reward and augment it with \gls{pbrs} based on our model~\cite{ng1999policy}.
\Gls{pbrs} suits the use case of a reward model from images particularly well, as it comes with a theoretical guarantee that the optimal policy remains unchanged, circumventing well-known failure modes of \gls{rl}~\cite{randlov1998learning}.
Let $\phi: S\to\mathbb{R}$ be a function of states $s\in S$ and $F: S\times A \times S \to \mathbb{R}$ with $F(s, a, s^\prime) = \gamma\phi(s^\prime) - \phi(s)$.
Then $F$ is called a \emph{potential-based} shaping function, and the optimal policy when replacing a reward $R$ with $R+F$ remains unchanged.
In our case, we utilize our model as the potential function $\phi$.
Note, however, that our reward model is not a function of state, but a function of a partial observation of state, and the \gls{pbrs} guarantees do not apply, as we will discuss in the next session.

Figure \ref{fig:rl} shows the fraction of runs solved after each evaluation.
We evaluate every 50,000 training steps and consider the task solved if the policy can solve it ten times in a row.
We compare our method with the analytical rewards of Meta-World+ and a sparse version, and show 95\% bootstrapped confidence intervals.
While some tasks are solved with sparse rewards (e.g.\@ window-close), most are only solved with dense or predicted shaped rewards, whereas assembly is not solved at all.

\subsection{Limitations}
A theoretical limitation of our model is that it is not Markov with respect to the true state, i.e.\@ the reward is not a function of state.
The reason is that an image is only a partial observation of the true state, so our model can assign the same reward to identical images when part of the state relevant to the task is not observable.
This also violates the theoretical guarantees of \gls{pbrs}, although in practice, in Meta-World+ and with our two-camera configurations, the entire scene is usually visible.
A reward model based on images can, however, still be theoretically sound, provided it has a memory unit or can observe the entire trajectory.
The reward is then Markov with respect to so-called belief states~\cite{kaelbling1998planning}.
Even the theoretical guarantees of \gls{pbrs} can then be recovered as shown by \mbox{Eck \etal{}~\cite{eck2016potential}}.

As discussed in Section~\ref{subsec:method-loss}, our loss is not defined on equal pairs, and our model is also not trained to detect successes.
While success detection is important, it is difficult to incorporate in a principled manner.
It usually implies equal rewards, which a model outputting real scalars can hardly achieve.
The auxiliary head approach proposed by Liang \etal{}~\cite{liang2026robometer} is a promising addition.
 
\section{Conclusion and Future Work}
In this paper, we present \method{}, a method for training dense reward functions from analytical rewards using pre-trained image and language representations.
Our results show that modern vision encoders contain sufficient information to recover precise reward signals.
The learned rewards are well-calibrated, generalize across prompt paraphrases and task variations, and enable dense reward prediction on real robots, despite the small dataset size.
Importantly, our method is simple and general enough to incorporate diverse datasets for reward learning.
It can also be used for \gls{rl} in a principled manner using \gls{pbrs}.
By evaluating against a ground truth and training policies from scratch, we show the effect of replacing analytical rewards with predicted imperfect rewards.

Our work can serve as a foundation for training reward models on larger datasets and with larger model capacities.
Scaling model capacity with sequence models, in particular, may improve the reliability of \gls{rl} algorithms by maintaining \gls{pbrs}'s theoretical guarantees.
Overall, we demonstrate that precise image-based reward models can be trained with compact architectures and datasets when using strong pretrained representations.

\section*{Acknowledgment}
This work has been partially supported by the Robotics Institute Germany (RIG) (grant no.~16ME1006) funded by the German Federal Ministry of Research, Technology and Space (BMFTR).
The authors acknowledge the HPC resources provided by the Erlangen National HPC Center (NHR@FAU) under the BayernKI project no.~v106be.

\IEEEtriggeratref{20}
\bibliography{bibliography.bib}
\end{document}